\documentclass[runningheads]{llncs}
\usepackage{multirow}
\usepackage[pdftex]{graphicx}
\usepackage{float}
\usepackage{hyperref}
\usepackage{multirow}
\usepackage{amsmath}
\usepackage{hyperref}
\usepackage{xcolor}
\usepackage{mathtools}
\usepackage{algorithm}
\usepackage{algpseudocode}

\begin{document}
\title{Information Gain Sampling for Active Learning in Medical Image Classification}  \titlerunning{Information Gain Active Learning}

\author{Raghav Mehta\inst{1,2} \and
Changjian Shui\inst{1,2} \and Brennan Nichyporuk\inst{1,2} \and Tal Arbel\inst{1,2} }

\authorrunning{R. Mehta et al.}

\institute{Centre for Intelligent Machines, McGill University, Montreal, Canada\\
\and MILA Quebec AI Institute, Montreal Canada}

\maketitle  

\begin{abstract}
Large, annotated datasets are not widely available in medical image analysis due to the prohibitive time, costs, and challenges associated with labelling large datasets. Unlabelled datasets are easier to obtain, and in many contexts, it would be feasible for an expert to provide labels for a small subset of images. This work presents an information-theoretic active learning framework that guides the optimal selection of images from the unlabelled pool to be labeled based on maximizing the expected information gain (EIG) on an evaluation dataset. Experiments are performed on two different medical image classification datasets: multi-class diabetic retinopathy disease scale classification and multi-class skin lesion classification. Results indicate that by adapting EIG to account for class-imbalances, our proposed Adapted Expected Information Gain (AEIG) outperforms several popular baselines including the diversity based CoreSet and uncertainty based maximum entropy sampling. Specifically, AEIG achieves $\sim 95\%$ of overall performance with only 19\% of the training data, while other active learning approaches require around 25\%. We show that, by careful design choices, our model can be integrated into existing deep learning classifiers.  

\end{abstract}

\section{Introduction}

The performance of deep learning methods is largely dependent on the availability of large, labelled datasets for model training~\cite{DataDep}. However, large, annotated datasets are not widely available in medical image analysis due to the prohibitive time, costs, and challenges associated with labelling large datasets. The labeling task is particularly challenging in patient images with pathological structures (e.g., lesions, tumours) and requires significant clinical and domain expertise. Various approaches have been proposed for optimally leveraging a small subset of annotated data that has been (passively) provided along with an otherwise unlabelled medical imaging dataset. These approaches range from transfer learning~\cite{SSTL,ImageNetTL}, weakly supervised~\cite{WSCV,WSMI}, semi-supervised \cite{SSCV,SSMI} to synthetic data generation~\cite{SDCV,SDMI}. \\

Active learning (AL) frameworks~\cite{ALLS,ALMICL}, on the other hand, have been successfully developed for "human-in-loop" computer vision~\cite{ALCVCL} and medical imaging classification contexts~\cite{ALMICL}. A comprehensive survey of active learning methods in medical image analysis can be found in~\cite{ALsurvey}. These AL approaches  work by training a model on a small, available, labeled subset, running inference on the larger unlabeled dataset, and then identifying an optimal set of samples to be labelled and added to the training pool in an iterative fashion. Sampling is optimized to attain the highest performance with the smallest number of samples. Sampling strategies can be broadly categorized as: (i) {\it uncertainty based}, which includes selecting samples with the least confidence in its estimated most probable class~\cite{LeastConf}, the smallest margin between the first and second most probable classes~\cite{SmallestMargin}, the maximum predicted entropy~\cite{Entropy}, the minimum expected generalization loss~\cite{EGL}, as well as deep Bayesian active learning approaches~\cite{DBAL} (MCD-Entr and MCD-BALD~\cite{BALD}) and (ii) {\it representative based}, which focuses on selecting the most representative and diverse images from the unlabeled set (e.g. CoreSet~\cite{CoreSet}, variational adversarial~\cite{VAAL,WAAL}, reinforcement learning~\cite{RLAL}). Combinations of multiple strategies~\cite{SuggestiveAnnotation,ConfCoreSet,LL} have also been proposed. \\

Generally, uncertainty-based active learning approaches, particularly entropy-based methods, have been popular in medical imaging contexts where they have shown some effectiveness in addressing the issue of high-class imbalance. %
However, while entropy based methods select the samples which are the hardest for the current model to classify, entropy alone does not convey the particular source of the uncertainty (e.g., which classes are the source of confusion in a multi-class classification task). In addition, it does not provide information about how the addition of the sample’s labels will influence downstream performance. \\

This paper proposes an information-theoretic active learning framework that drives the selection of new image samples to label based on maximal information gain. An active learning framework which selects samples based on Expected Information Gain (EIG) has been previously used~\cite{EGL} for structure prediction tasks using Support Vector Machines (SVM). As the first contribution of this paper, we first adapt an efficient EIG computation to deep networks with careful design choices. In order to alleviate the high class-imbalance issue in medical imaging, we further improve the original EIG by proposing a novel Adapted Expected Information Gain (AEIG) method. In AEIG, the predicted softmax probability of the trained model is adjusted with the class frequencies of the validation distribution. The hypothesis is that AEIG based sampling strategy will lead to higher performance with a lower number of labeled samples, as different labelled samples provide different information about inter-class ambiguity. \\

Experiments are performed on two different challenging medical image classification tasks: (1) multi-class diabetic retinopathy (DR) classification into disease scales from colour fundus images, (2) multi-class skin lesion classification from dermoscopic images. Our experiments indicate that for the DR dataset AEIG achieves 95\% of overall performance with only 19\% of the training data. In comparison, other active learning methods require around 25\% (random: 27\%, maximum entropy: 21\%, CoreSet: 27\%, MCD-Entropy: 24\%, MCD-BALD: 21\%). AEIG achieves higher performance than competing methods due to its ability to sample more images from the minority classes compared to other methods on highly imbalanced datasets.

\begin{figure*}[!t]
\centering
\includegraphics[width=1.00\textwidth]{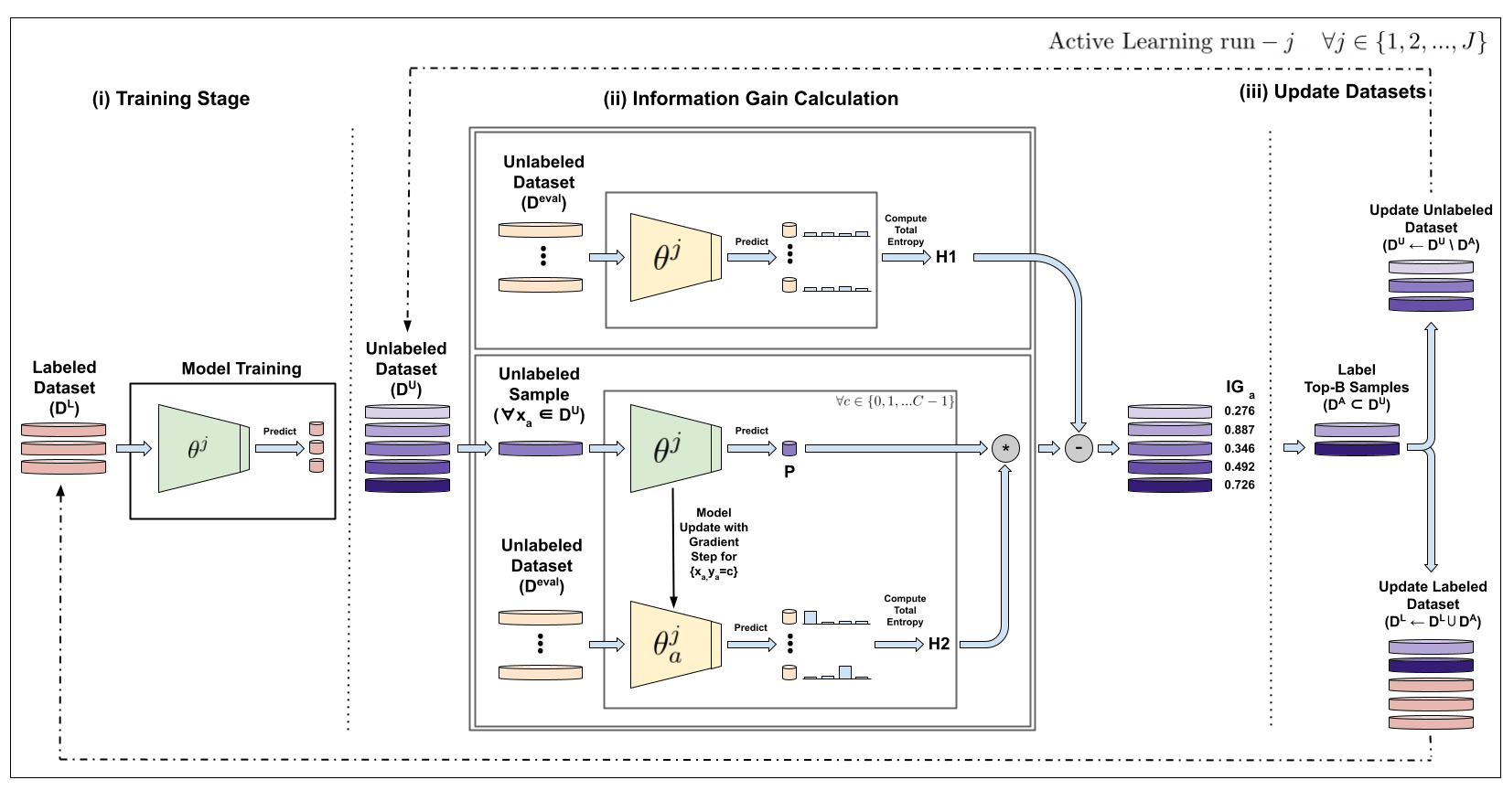}
\center \caption{Active Learning via Information Gain  framework. Each active learning run consists of three different phases: \textbf{(i) Training Stage} - Model ($\theta^{j-1} \rightarrow \theta^j$) is trained using the labeled set $D^L$, \textbf{(ii) Information Gain Calculation} - $\text{AEIG}_a$ (Equation\eqref{AEIG}), $\text{EIG}_a$ (Equation\eqref{EIG}), or its variants are calculated for each image in the unlabeled dataset ($ \forall x_a \in D^U$). The entropy H1 of the evaluation set ($D^{\text{eval}}$) is calculated using the trained model ($\theta^j$). For each image $x_a$, The conditional entropy (H2) of the evaluation set is calculated after updating the trained model ($\theta^j$) using a single gradient step ($\theta^j \rightarrow \theta_a^j$) for all possible labels $y_a=c$, $\forall c \in \{0,1,...,C-1\}$.  \textbf{(iii) Update Datasets} - Finally, the top-B images ($D^A$) from the unlabeled set is selected and both the labeled ($D^L \gets D^L \cup D^A$) and unlabeled datasets ($D^U \gets D^U \setminus D^A$) are updated. The framework is executed for a total of $J$ runs.}
\label{fig:EIG}
\end{figure*}

\section{Active Learning Framework with Information Gain Sampling}

Consider a labeled training dataset $D^L: \{(X_L,Y_L)\}$. Here, $(X_L,Y_L) = \{(x_i,y_i=c)\}_{i=1}^{M}$, represents that there are a total of M samples ($x_i$) in the dataset; and $y_i=c$ represents its corresponding classification label, where there are a total of $C$ classes ($c \in \{0,1,...,C-1\}$). Now, consider an unlabeled dataset $D^U: \{(X_U)\}$, with N samples. Similarly, an evaluation dataset $D^{\text{eval}}: \{(X_{\text{eval}},Y_{\text{eval}})\}$ with K samples. Here, $X^{\text{eval}}$ represents the set of all samples in the evaluation set, and $Y^{\text{eval}}$ its corresponding labels.  $\hat{Y}^{\text{eval}}$ would represent the predicted classification label for each sample in the evaluation set using a machine learning model. Note that $M \ll N$ and $K < N$. \\

The general active learning framework starts by training a supervised machine learning model ($\theta^0$) on a small labeled dataset ($D^L$). It then selects the $B$ most informative subset of images to label ($D^A: \{x_a\}_{a=0}^{B}$, $D^A \subset D^U$) from a larger unlabeled dataset ($D^U$). A human-annotator provides the labels for the selected subset of data ($D^{A*}: \{(X_A, Y_A\} = \{x_a,y_a\}_{a=0}^{B}$). Both the labeled ($D^L \gets D^L \cup D^{A*}$) and the unlabeled datasets ($D^U \gets  D^U \setminus D^A$) are then updated. The model is retrained using the updated labeled dataset ($\theta^0 \rightarrow \theta^1$). The process is repeated for a total of $J$ runs.

\subsection{Information Gain (IG) for Active Learning}

An active learning framework can select the subset of data from the unlabeled set based on the information gain. \\

\textbf{Expected Information Gain (EIG):} Let us consider the case of Expected Information Gain (EIG). In active learning context, EIG$(\hat{Y}^{\text{eval}}; y_a)$ measure the reduction in the entropy of the predicted labels $\hat{Y}^{\text{eval}}$ of the evaluation set, if we have access to the true state (label - $y_a$) of an image ($x_a$) in the unlabeled set. In short, EIG$(\hat{Y}^{\text{eval}}; y_a)$ measures difference in the entropy of $\hat{Y}^{\text{eval}}$ for two models. (i) \textbf{H1}: the entropy of the $\hat{Y}^{\text{eval}}$ for a model trained on $D^L$.  (ii) \textbf{H2}: the conditional entropy of the $\hat{Y}^{\text{eval}}$ for a model trained on $\{D^L \cup (x_a, y_a)\}$. \\

\vspace{-3mm}
\begin{equation}
    \begin{split}
        & \text{EIG}(\hat{Y}^{\text{eval}}; y_a) = {\text{EIG}} (\hat{Y}^{\text{eval}}; y_a |  X^{\text{eval}}, x_a, D^{L}) \\
        & = {\textbf{H}} [\hat{Y}^{\text{eval}} | X^{\text{eval}}, D^{L}] - {\textbf{H}} [\hat{Y}^{\text{eval}} | X^{\text{eval}}, y_a, x_a, D^{L}]  \\
        & = \underbrace{{\textbf{H}} [\hat{Y}^{\text{eval}} | X^{\text{eval}}, D^{L}]}_{\textbf{H1}} - \sum_{c=0}^{C-1} \underbrace{p(y_a=c | x_a, D^L)}_{\textbf{P}} \underbrace{ {\textbf{H}} [\hat{Y}^{\text{eval}} | X^{\text{eval}}, y_a=c, x_a, D^{L}] }_{\textbf{H2}} \\        
    \end{split}
\label{EIG}
\end{equation}

$\textbf{P} = {p}(y_a=c | x_a, D^L)$ denotes the predicted softmax probability of output having class label $y_a=c$ for an image $x_a$ using a model trained on $D^L$.  \\

\textbf{Adapted Expected Information Gain (AEIG):} The predicted softmax probability $\textbf{P}$ can be quite erroneous, due to the limited observations and poor calibration \cite{guo2017calibration}. Thus, other alternatives can be considered to improve the reliability of $\textbf{P}$ such as injecting prior information about the class distribution. In the natural image classification literature, several methods \cite{Adjust1,Adjust2,Adjust3} have been proposed that adadpt the softmax probabilities in the context of highly imbalanced datasets. As such, a variant of the EIG method is considered here, where the predicted softmax probability (\textbf{P}) of the training model is adjusted with the class frequencies of the validation distribution.  The adapted version of EIG, denoted Adapted Expected Information Gain (AEIG), provides a modification for $\textbf{P}$ to become $\text{p}(y_a=c | x_a, D^L) * \frac{|y_{\text{eval}}=c|}{\sum_{j=0}^{C-1} |y_{\text{eval}}=j|}$, where $|y_{\text{eval}}=c|$ denotes the total number of samples with class-label $c$ in the evaluation dataset:
\begin{equation}
    \begin{split}
        & \text{AEIG}(\hat{Y}^{\text{eval}}; y_a) = {\textbf{H1}} - \underbrace{p(y_a=c | x_a, D^L)\frac{|y_{\text{eval}}=c|}{\sum_{j=0}^{C-1} |y_{\text{eval}}=j|}}_{\textbf{P}} {\textbf{H2}}.         
    \end{split}
\label{AEIG}    
\end{equation}

\subsection{Efficient IG computation in Deep Networks}

As we saw in the previous section, computing both EIG~\eqref{EIG} and AEIG~\eqref{AEIG} involes estimating the conditional entropy (\textbf{H2}) by retraining the models for each possible label for an image (i.e., a total of $C$ classes) in the unlabeled set. In the active learning framework, this calculation is repeated for each image in the unlabeled set (i.e., total N images). Although this process might be feasible in the context of SVMs~\cite{EGL}, it would be very computationally expensive (almost infeasible) in the context of a deep learning model (i.e., a total N*C model retraining). In this paper, design simplifications are made to reduce the associated computation load. \\

\noindent\textbf{Choice of Evaluation Set:} In the first design simplification, we consider the validation set as our evaluation dataset ($D^{\text{eval}} = D^{\text{valid}}$). \\

\noindent\textbf{Model Parameters:} The second design simplification is based on the observation \cite{PC} that any machine learning model, including deep learning, consists of two components: representation and classification. In the context of modern convolutional neural network architectures, initial convoltional layers can be considered as a feature representation learning layers, while the last MLP layers can be considered as a classification layer. While updating the model parameters during the IG calculation, only the classification layer parameters are updated. The convolutional layer's parameters are not updated. Given that most of the computation cost comes from the convolutional layers, this design permits computing IG scores (EIG or AEIG) with minimal computational overhead. \\

\noindent\textbf{Model Updates:} In the third design simplification, instead of retraining the whole model with the labeled dataset and each sample in the unlabeled dataset, the already trained model on labeled set is only updated once through a single gradient step for one sample in the unlabeled set. This design simplification is based on the assumption that the size of labeled dataset is greater than a single sample, and inclusion of just one sample would not lead to a drastic change in the model parameters.

\section{Multi-Class Medical Image Disease Classification}
The active learning framework is applied to two different medical imaging contexts. The first context involves multi-class disease classification of Diabetic Retinopathy (DR) patients from colour fundus images. Fundus images are classified into five disease scales representing disease severity: No DR, Mild DR, Moderate DR, Severe DR, Proliferative DR. A publicly available DR disease scale classification dataset is used for this task. This paper uses a subset of 8408 retinal fundus images provided by the kaggle challenge organizers. A label with one of the five disease scales is provided with each retinal fundus image. For each of the five disease scales there are 6150/588/1283/221/166 images, respectively. The differences in the total number of images per class highlight a high-class imbalance for the task. We randomly divide the whole dataset into 5000/1000/2408 images for training/validation/testing sets.\\

The second context involves multi-class classification of skin lesions from dermoscopic images. We use publicly available International Skin Imaging Collaboration (ISIC) 2018 dataset \cite{ISIC18}. In this dataset, dermoscopic images are classified into 7 different classes: Melanoma, Melanocytic nevus, Basal cell carcinoma, Actinic keratosis, Benign keratosis, Dermatofibroma, and Vascular lesion. The challenge organizers provide a subset of 10015 dermoscopic images. A label with one of the seven disease scales is provided with each dermoscopic image. For each of the seven classes there are 1113/6705/514/327/1099/115/142 images, respectively. The differences in the total number of images per class highlight a high-class imbalance for the task. We randomly divide the whole dataset into 6000/1500/2515 images for training/validation/testing sets.

\section{Experiments and Results}
\subsection{Implementation Details}
\noindent \textbf{Network Architectures:} 
For both tasks, the DR and the ISIC multi-class disease scale classification, an imagenet pre-trained ResNet-18 architecture~\cite{ResNet} was deployed. A Dropout layer with p=0.2 was introduced before the MLP layer. The networks were trained for a total of 100 epochs, using the Adam optimizer with a learning rate of 0.0001/0.0005 for ISIC/DR datasets ~\footnote{The exact architecture and training details are provided in the appendix.}~\footnote{Code: \url{https://github.com/RagMeh11/IGAL}}. The 'macro' Area Under the Receiver Operating Characteristic Curve (ROC AUC) was used as a metric for both classification tasks. For both tasks, a macro average (unweighted) of one-vs-rest (ovr) classifier ROC AUC~\cite{ROC-AUC} was performed. \\

\noindent \textbf{AL framework:} The active learning framework was initialized by randomly selecting 10\% of the training dataset (i.e., 500 for DR, 600 for ISIC) as the labeled dataset and the rest as the unlabeled dataset. It was deployed for a total of $J=6$ active learning runs in both cases. Based on previous studies \cite{RLAL,ConfCoreSet}, in each run, we select a total of $\approx$ 6\% of the dataset ($B$ = 300 for the DR, and $B$ = 350 for the ISIC)  from the unlabeled dataset ($D^U$). We acquire an oracle label, and then, once labelled, these are used to update the labeled dataset ($D^L \gets D^L \cup D^{A*}$) and the unlabeled dataset ($D^U \gets D^U \setminus D^{A}$). Active learning experiments were repeated five times with different initial randomly selected images. The means and variances of the evaluation metrics were then recorded across the five repetitions.

\begin{figure*}[!t]
\centering
\includegraphics[width=1.00\textwidth]{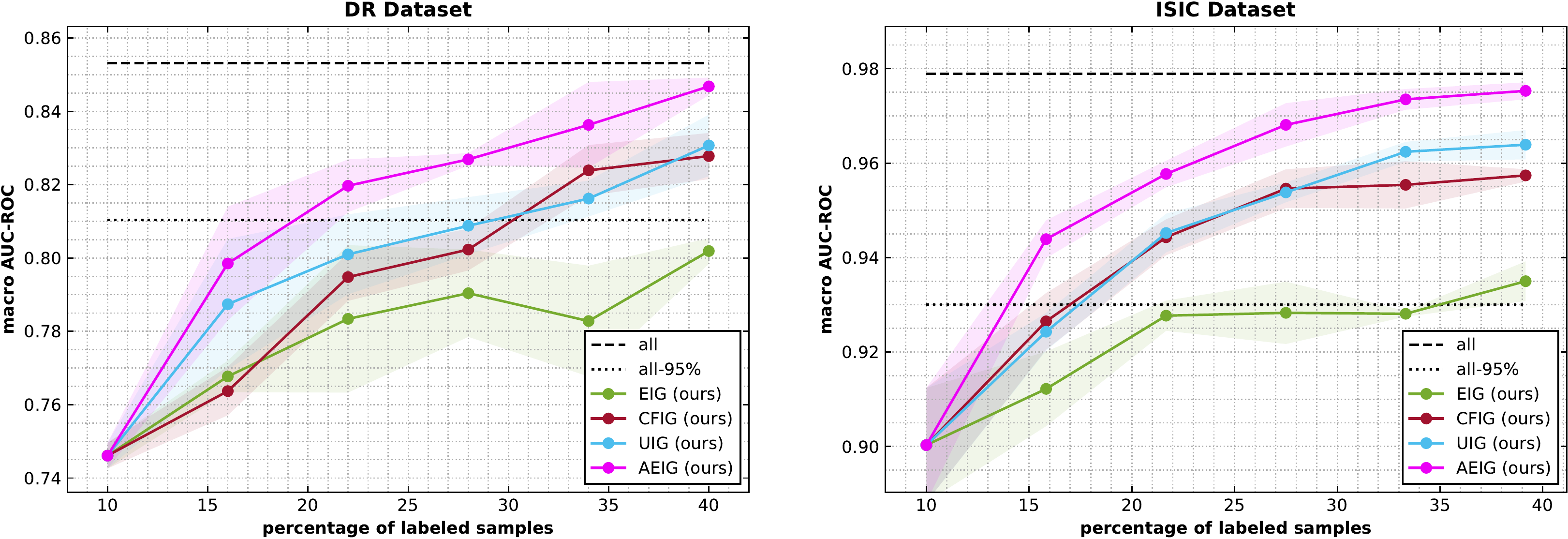}
\center \caption{Comparison of the EIG, AEIG, UIG, and CFIG based active learning sampling methods for both the DR dataset (left) and the ISIC dataset (right). The horizontal solid dashed line ('all') at the top represents model performance with the entire training set is labeled. The doted line ('all-95\%') represents 95\% of that performance. We report mean and std of evaluation metric across five different runs. (See Table-1 and Table-2 in the appendix for exact values.)}
\label{fig:IG_comparison}
\end{figure*}

\subsection{Information Gain Performance}
In this section, we compare the proposed AEIG and EIG based active learning sample selection against two different baseline alternatives for IG computation. Equation~\eqref{EIG} describes the estimation of EIG, which includes weighing H2 with the predicted softmax probability \textbf{P}. Instead of relying on the predicted probabilities, we can compute two different baseline alternatives based on the prior information of the class distributions: (i) Uniform Information Gain (UIG) assumes a uniform distribution such that $\textbf{P} = \frac{1}{C}$, $\forall c \in \{0,1,...,C-1\}$.  (ii) Class-Frequency Information Gain (CFIG) assumes a distribution based on the class frequency such that $\textbf{P} = \frac{|y_{\text{eval}}=c|}{\sum_{j=0}^{C-1} |y_{\text{eval}}=j|}$, where $|y_{\text{eval}}=c|$ denotes the total number of samples with class-label $c$ in the evaluation dataset.\\

In Figure~\ref{fig:IG_comparison}, we compare EIG, UIG, CFIG, and AEIG by experimenting on both datasets. Experiments indicate that the AEIG achieves 95\% of the overall performance ('all-95\%') with only 19\% (for DR) and 14\% (for ISIC) of the training dataset. CFIG, UIG, and EIG requires 29\%, 30\% and $>$40\% of the training dataset for DR; and 17\%, 17.5\%, and 35\%  of the training dataset for ISIC. We hypothesize that better performance of AEIG is due to its ability of sampling more images from minority classes \footnote{See Fig:2 and Fig:3 in the appendix.}.

\begin{figure*}[!t]
\centering
\includegraphics[width=1.00\textwidth]{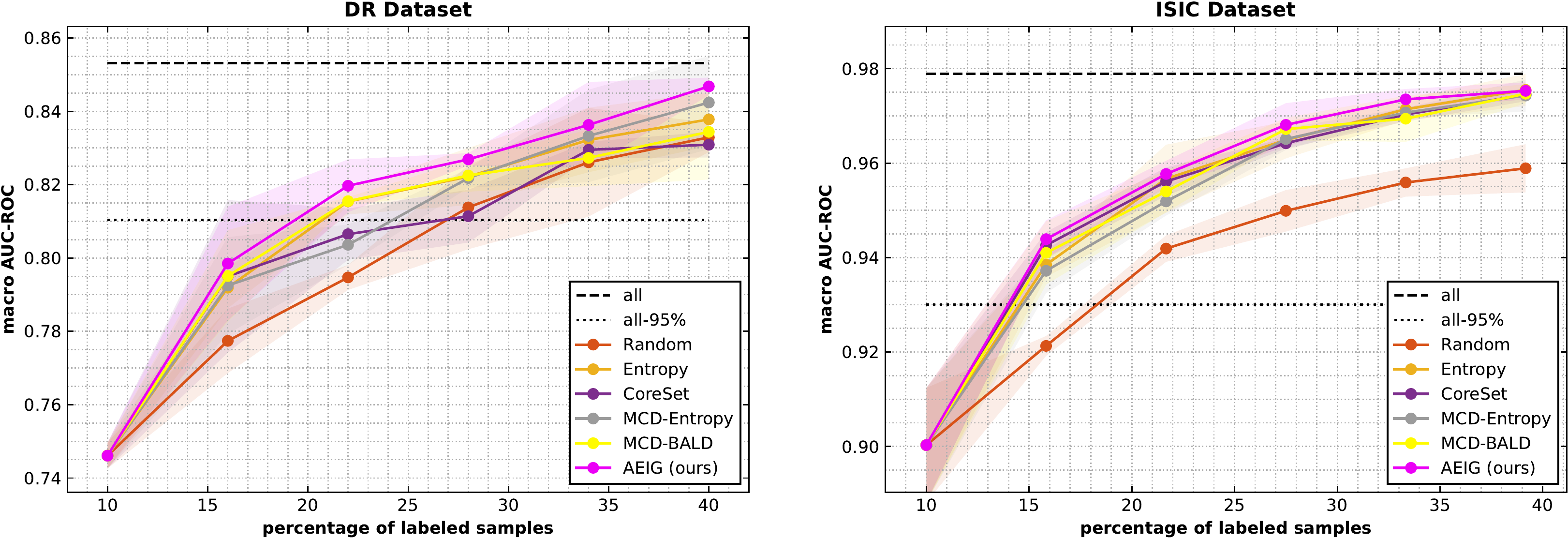}
\center \caption{Comparison of the AEIG based active learning sampling method with Random, Entropy, MCD-Entropy, MCD-BALD, and CoreSet based sampling methods for both the DR dataset (left) and the ISIC dataset (right). The horizontal solid dashed line ('all') at the top represents model performance with the entire training set is labeled. The doted line ('all-95\%') represents 95\% of that performance. We report mean and std of evaluation metric across five different runs. (See Table-3 and Table-3 in the appendix for exact values.)}
\label{fig:AL_comparison}
\end{figure*}

\subsection{Comparisons Against Active Learning Baselines}
In this section, the proposed AEIG based sampling active learning framework was compared against five different baseline methods: Random, Entropy-based sampling~\cite{Entropy}, MC-Dropout with Entropy \cite{DBAL}, MC-Dropout with BALD \cite{BALD}, and CoreSet~\cite{CoreSet}. The macro AUC ROC curve for experiments on the DR and ISIC datasets can be found in Figure~\ref{fig:AL_comparison}. Overall, the proposed method gives better (or in some cases similar) performance compared to the other methods for both datasets and all six active learning iterations. Applying standard methods for comparison, the proposed method (AEIG) achieves 95\% of the overall performance (’all-95\%’) with only 19\% of the labeled training dataset for the DR dataset. MCD-Entropy, MCD-BALD, Entropy, CoreSet, and Random require approximately 24\%, 21\%, 21\%, 27\%, and 27\% of the labeled training dataset to achieve similar performances. For the ISIC dataset, the proposed method (AEIG) achieves 95\% of the overall performance (’all-95\%’) with only 14\% of the labeled training dataset for the DR dataset. MCD-Entropy, MCD-BALD, Entropy, CoreSet, and Random require approximately 14.8\%, 14.2\%, 14.7\%, 14.1\%, and 18.2\% of the labeled training dataset to achieve similar performances. It is worth pointing out although all methods are giving somewhat similar performance at 'all-95\%' cutoff, the trend is consistant for all 6 AL acquisitions.  The total active learning score computational time for each image in the unlabeled set was around 1 ms, 6 ms, 10 ms, 10 ms, 16 ms, and 28 ms for Random, Entropy, MCD-Entropy, MCD-BALD, CoreSet, and AEIG based methods. The computation times highlight that although the proposed method can achieve better performance in comparison to other methods, it is a bit slower. Compared to the time taken by a human annotator for additional labelling, this difference in computational time will not be significant. 

\begin{figure*}[!t]
\centering
\includegraphics[width=1.00\textwidth]{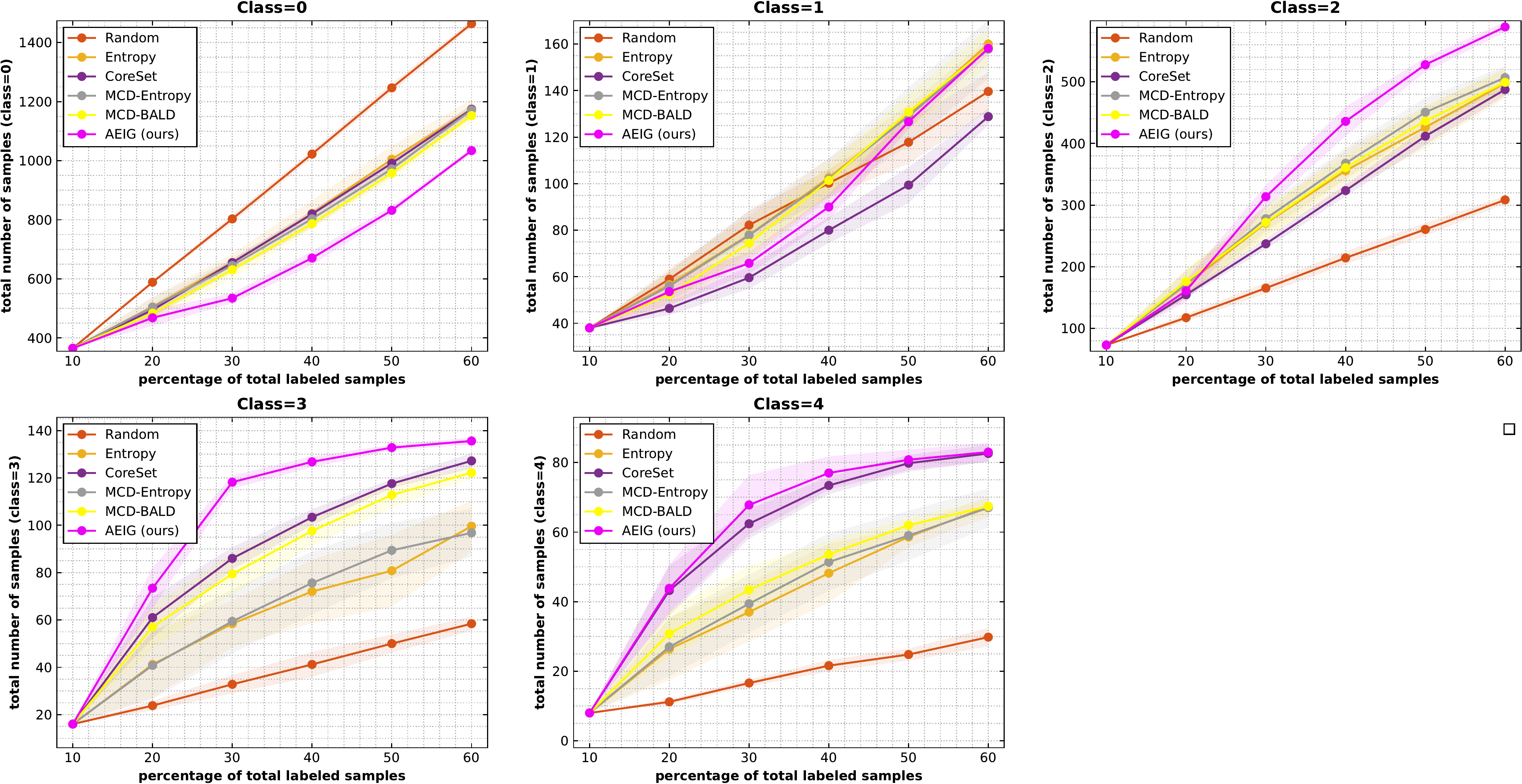}
\center \caption{Plots depicting the total number of samples labelled per class against the percentage of labeled samples for the DR dataset for the competing active learning sampling methods. Classes 1,3, and 4 are the minority classes. }
\label{fig:Acqu_comparison}
\end{figure*}

Figure~\ref{fig:Acqu_comparison} illustrates the different number of acquired images per class on the DR dataset at each of the active learning acquisition steps for all six acquisition methods (Random, Entropy, CoreSet, MCD-Entropy, MCD-BALD, and AEIG). The results indicate that the AEIG based active learning sampling policy results in sampling and labelling of a higher number of images from the minority classes (e.g., classes 1, 3, and 4) compared to other sampling methods. This, in turn, leads to better overall performance for contexts with highly class-imbalance datasets, as is the case with the DR dataset~\footnote{Similar curves for the ISIC dataset are included in the appendix.}.

\section{Conclusions}
In this paper, we proposed an active learning framework that drives the selection of new image samples to label based on maximal Adapted expected information gain on an unseen evaluation dataset. Experiments were performed on two different medical image classification datasets, and results showed that the AEIG method achieves better performance than Random, maximum Entropy, MCD-Entropy. MCD-BALD, and CoreSet based sampling strategies. The AEIG samples minority classes at a greater rate than competing strategies, improving performance on highly imbalanced datasets, although with a small computational overhead. \\

\noindent\textbf{Acknowledgement:} This investigation was supported by the Natural Sciences and Engineering Research Council of Canada, the Canada Institute for Advanced Research (CIFAR) Artificial Intelligence Chairs program, and a technology transfer grant from Mila - Quebec AI Institute.

\section{Supplementary Material}

\subsection*{Algorithm}

\begin{algorithm}
\caption{Information Gain Based Active Learning}
\hspace*{\algorithmicindent} \textbf{Input:} Labeled training dataset  $D^L: \{(X^L,Y^L)\}$, an unlabeled dataset $D^U: \{ (X^U) \}$, and an evaluation dataset $D^{eval}$ \\
\begin{algorithmic}[1]
\Require initial machine model (with parameters $\theta^0$) trained on labeled dataset $D^L$, total active learning iterations $J$, and active learning batch acquisition size $B$ 
\State $j \gets 1$
\While{active learning iteration $j < J$} \\
    \State Calculate ${\textbf{H}} [\hat{Y}^{\text{eval}} | X^{\text{eval}}, D^{L}]$ based on the model parameters $\theta^{j-1}$  \\
    \For{each image $x_a \in D^U$} 
        \State Calculate ${\text{p}}(y_a = c | x_a, D^{L})$ based on the model parameters $\theta^{j-1}$
        \State $\theta^{j-1}_a \gets \theta^{j-1}$ \\
        \For{each possible class label $c \in \{0,1,..,C\}$}
            \State Using a single gradient step update model parameters ($\theta^{j-1}_a$) with $x_a$ and $y_a=c$
            \State Calculate ${\textbf{H}} [\hat{Y}^{\text{eval}} | X^{\text{eval}}, y_a=c, x_a,  D^{L}]$
        \EndFor \\
        \State Compute Score based on AEIG (or EIG) according to Equation [2] (or [1])
    \EndFor \\
    \State Select subset of top-B images ($D^A$) from $D^U$ according to their score $S$
    \State Acquire ground-truth labels for $D^A$ (($D^{A*}$))
    \State Update Unlabeled dataset $D^U \gets D^U \setminus D^A$
    \State Update Labeled dataset $D^L \gets D^L \cup D^{A*}$
    \State Retrain the model ($\theta^j$) with the updated labeled training dataset $D^L$
    \State $j \gets j + 1$ \\
\EndWhile
\end{algorithmic}
\end{algorithm}

\begin{figure*}[!t]
\centering
\includegraphics[width=1.00\textwidth]{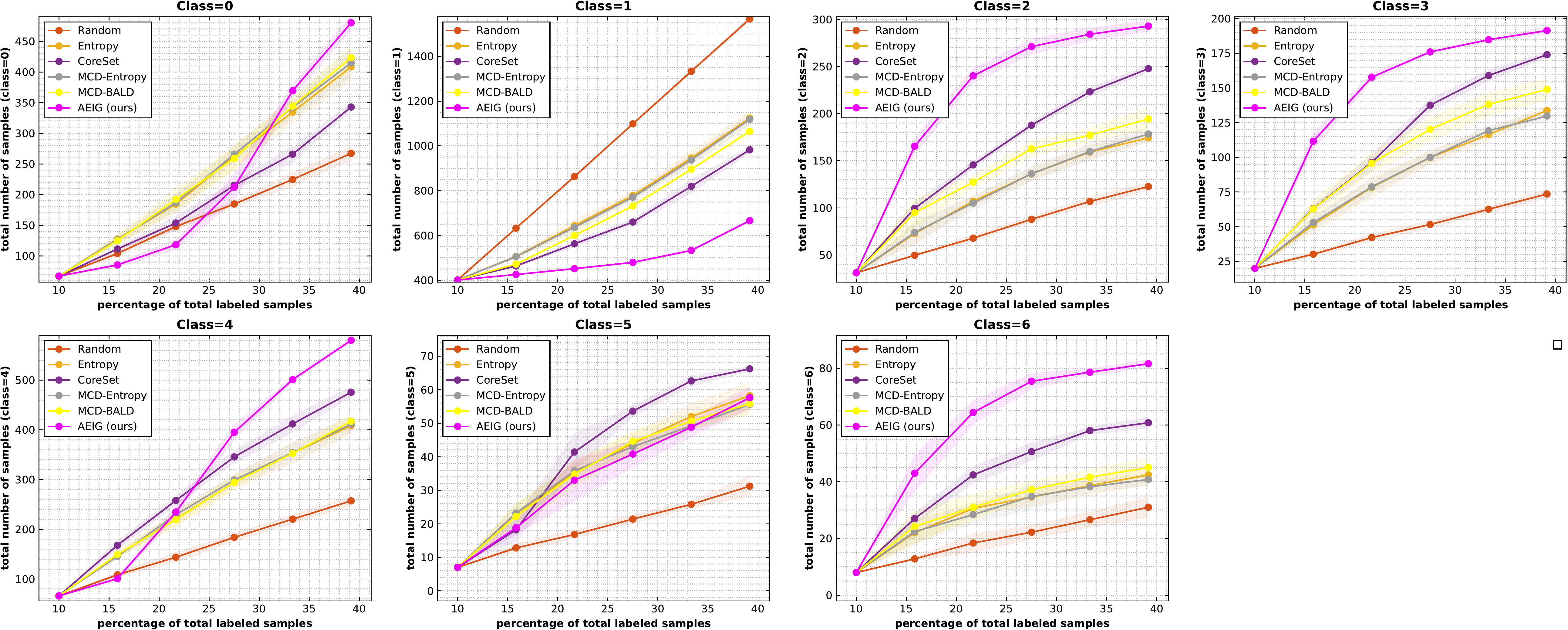}
\center \caption{Plots depicting the total number of samples labelled per class against the percentage of labeled samples for the ISIC dataset for the competing active learning sampling methods. Classes 0, 2,3,4,5, and 6 are the minority classes. }
\label{fig:Acqu_comparison_ISIC}
\end{figure*}

\begin{figure*}[!t]
\centering
\includegraphics[width=1.00\textwidth]{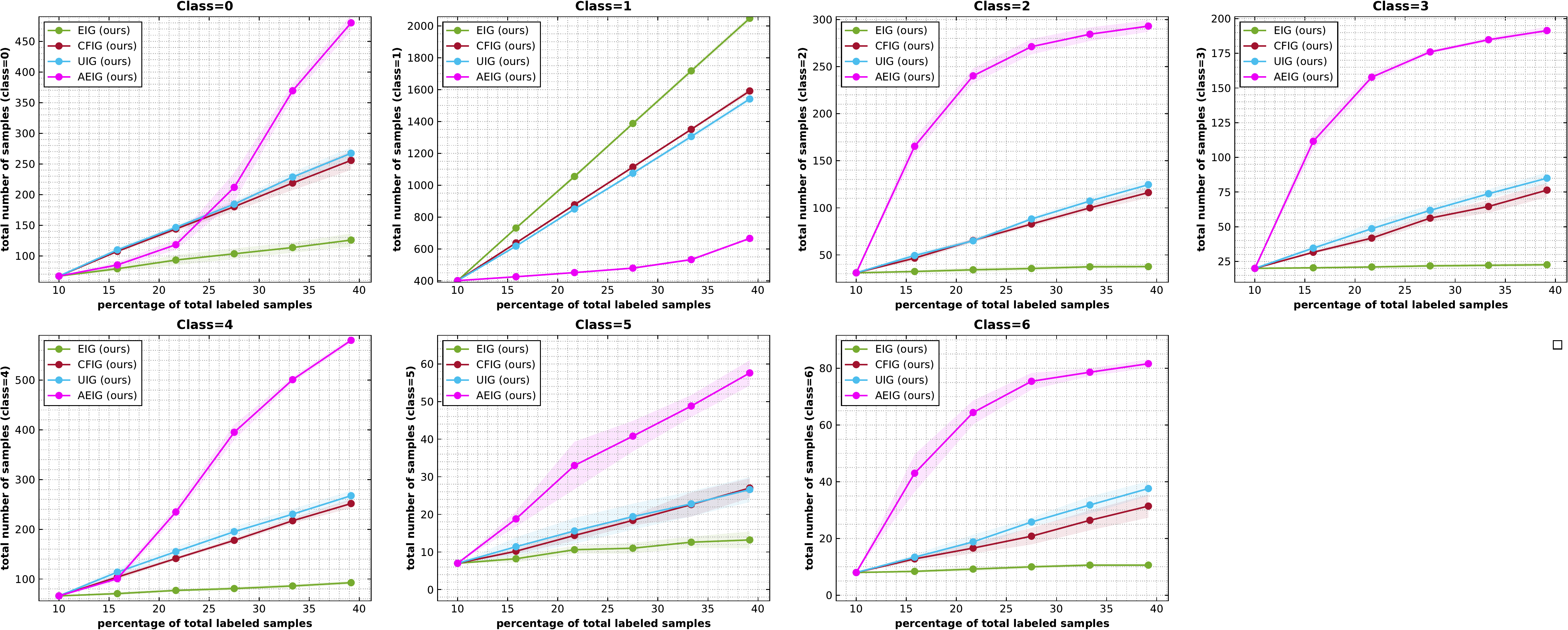}
\center \caption{Plots depicting the total number of samples labelled per class against the percentage of labeled samples for the ISIC dataset for EIG, CFIG, UIG, and AEIG sampling methods. Classes 0, 2,3,4,5, and 6 are the minority classes. }
\label{fig:Acqu_comparison_ISIC_IG}
\end{figure*}

\begin{figure*}[!t]
\centering
\includegraphics[width=1.00\textwidth]{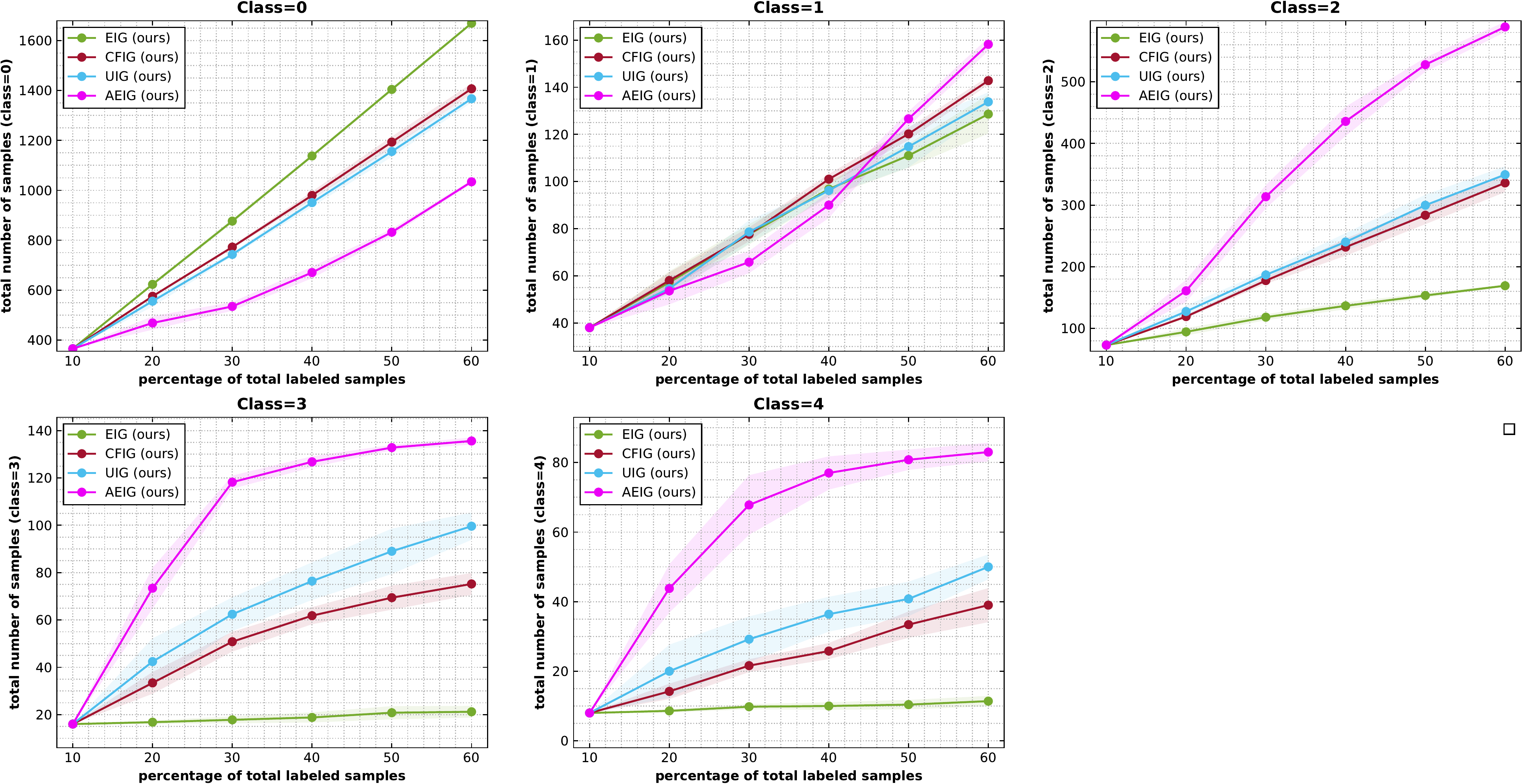}
\center \caption{Plots depicting the total number of samples labelled per class against the percentage of labeled samples for the DR dataset for EIG, CFIG, UIG, and AEIG sampling methods. Classes 1,3, and 4 are the minority classes. }
\label{fig:Acqu_comparison_DR_IG}
\end{figure*}

\begin{table}[]
\resizebox{\textwidth}{!}{%
\begin{tabular}{l||cccccc}
\hline \hline
\multirow{2}{*}{\textbf{Method}} & \multicolumn{6}{c}{\textbf{Percentage of Labeled Sample}}                                                                                                                                                                     \\ \cline{2-7} 
                                 & \multicolumn{1}{c|}{\textbf{10}}       & \multicolumn{1}{c|}{\textbf{16}}       & \multicolumn{1}{c|}{\textbf{22}}       & \multicolumn{1}{c|}{\textbf{28}}       & \multicolumn{1}{c|}{\textbf{34}}       & \textbf{40}       \\ \hline \hline
\textbf{EIG}                     & \multicolumn{1}{c|}{0.7461 $\pm$ 0.0035} & \multicolumn{1}{c|}{0.7677 $\pm$ 0.0043} & \multicolumn{1}{c|}{0.7834 $\pm$ 0.0201} & \multicolumn{1}{c|}{0.7904 $\pm$ 0.0119} & \multicolumn{1}{c|}{0.7828 $\pm$ 0.0151} & 0.8019 $\pm$ 0.0035 \\ 
\textbf{UIG}                     & \multicolumn{1}{c|}{0.7461 $\pm$ 0.0035} & \multicolumn{1}{c|}{0.7874 $\pm$ 0.0178} & \multicolumn{1}{c|}{0.8010 $\pm$ 0.0109} & \multicolumn{1}{c|}{0.8088 $\pm$ 0.0078} & \multicolumn{1}{c|}{0.8162 $\pm$ 0.0049} & 0.8307 $\pm$ 0.0084 \\ 
\textbf{PIG}                     & \multicolumn{1}{c|}{0.7461 $\pm$ 0.0035} & \multicolumn{1}{c|}{0.7637 $\pm$ 0.0066} & \multicolumn{1}{c|}{0.7948 $\pm$ 0.0065} & \multicolumn{1}{c|}{0.8023 $\pm$ 0.0057} & \multicolumn{1}{c|}{0.8239 $\pm$ 0.0069} & 0.8278 $\pm$ 0.0063 \\ 
\textbf{AEIG}                    & \multicolumn{1}{c|}{0.7461 $\pm$ 0.0035} & \multicolumn{1}{c|}{0.7985 $\pm$ 0.0155} & \multicolumn{1}{c|}{0.8197 $\pm$ 0.0072} & \multicolumn{1}{c|}{0.8269 $\pm$ 0.0017} & \multicolumn{1}{c|}{0.8363 $\pm$ 0.0117} & 0.8468 $\pm$ 0.0024 \\ \hline \hline
\end{tabular}%
}
\caption{Comparison of the EIG, AEIG, UIG, and CFIG based active learning sampling methods for both the DR dataset We report the mean and std of evaluation metric across five different runs. Model performance with the entire training set is 0.8561.}
\label{tab:DR-IG}
\end{table}

\begin{table}[]
\resizebox{\textwidth}{!}{%
\begin{tabular}{l||cccccc}
\hline \hline
\multirow{2}{*}{\textbf{Method}} & \multicolumn{6}{c}{\textbf{Percentage of Labeled Sample}}                                                                                                                                                                     \\ \cline{2-7} 
                                 & \multicolumn{1}{c|}{\textbf{10}}       & \multicolumn{1}{c|}{\textbf{15.83}}       & \multicolumn{1}{c|}{\textbf{21.67}}       & \multicolumn{1}{c|}{\textbf{27.5}}       & \multicolumn{1}{c|}{\textbf{33.33}}       & \textbf{39.17}       \\ \hline \hline
\textbf{EIG}                     & \multicolumn{1}{c|}{0.9033 $\pm$ 0.0121} & \multicolumn{1}{c|}{0.9122 $\pm$ 0.0079} & \multicolumn{1}{c|}{0.9277 $\pm$ 0.0032} & \multicolumn{1}{c|}{0.9283 $\pm$ 0.0066} & \multicolumn{1}{c|}{0.9281 $\pm$ 0.0006} & 0.9350 $\pm$ 0.0042 \\ 
\textbf{UIG}                     & \multicolumn{1}{c|}{0.9003 $\pm$ 0.0121} & \multicolumn{1}{c|}{0.9243 $\pm$ 0.0040} & \multicolumn{1}{c|}{0.9452 $\pm$ 0.0041} & \multicolumn{1}{c|}{0.9538 $\pm$ 0.0021} & \multicolumn{1}{c|}{0.9624 $\pm$ 0.0021} & 0.9639 $\pm$ 0.0031 \\ 
\textbf{PIG}                     & \multicolumn{1}{c|}{0.9003 $\pm$ 0.0121} & \multicolumn{1}{c|}{0.9265 $\pm$ 0.0060} & \multicolumn{1}{c|}{0.9443 $\pm$ 0.0038} & \multicolumn{1}{c|}{0.9546 $\pm$ 0.0041} & \multicolumn{1}{c|}{0.9554 $\pm$ 0.0050} & 0.9574 $\pm$ 0.0013 \\ 
\textbf{AEIG}                    & \multicolumn{1}{c|}{0.9003 $\pm$ 0.0121} & \multicolumn{1}{c|}{0.9439 $\pm$ 0.0040} & \multicolumn{1}{c|}{0.9577 $\pm$ 0.0028} & \multicolumn{1}{c|}{0.9681 $\pm$ 0.0046} & \multicolumn{1}{c|}{0.9735 $\pm$ 0.0022} & 0.9753 $\pm$ 0.0018 \\ \hline \hline
\end{tabular}%
}
\caption{Comparison of the EIG, AEIG, UIG, and CFIG based active learning sampling methods for both the ISIC dataset We report the mean and std of evaluation metric across five different runs. Model performance with the entire training set is 0.9789.}
\label{tab:ISIC-IG}
\end{table}

\begin{table}[]
\resizebox{\textwidth}{!}{%
\begin{tabular}{l||cccccc}
\hline \hline
\multirow{2}{*}{\textbf{Method}} & \multicolumn{6}{c|}{\textbf{Percentage of Labeled Sample}}                                                                                                                                                                     \\ \cline{2-7} 
                                 & \multicolumn{1}{c|}{\textbf{10}}       & \multicolumn{1}{c|}{\textbf{16}}       & \multicolumn{1}{c|}{\textbf{22}}       & \multicolumn{1}{c|}{\textbf{28}}       & \multicolumn{1}{c|}{\textbf{34}}       & \textbf{40}       \\ \hline \hline
\textbf{Random}                  & \multicolumn{1}{c|}{0.7461 $\pm$ 0.0035} & \multicolumn{1}{c|}{0.7774 $\pm$ 0.0088} & \multicolumn{1}{c|}{0.7947 $\pm$ 0.0034} & \multicolumn{1}{c|}{0.8138 $\pm$ 0.0114} & \multicolumn{1}{c|}{0.8261 $\pm$ 0.0148} & 0.8329 $\pm$ 0.0040 \\ 
\textbf{Entropy}                 & \multicolumn{1}{c|}{0.7461 $\pm$ 0.0035} & \multicolumn{1}{c|}{0.7919 $\pm$ 0.0056} & \multicolumn{1}{c|}{0.8154 $\pm$ 0.0035} & \multicolumn{1}{c|}{0.8222 $\pm$ 0.0078} & \multicolumn{1}{c|}{0.8322 $\pm$ 0.0088} & 0.8378 $\pm$ 0.0073 \\ 
\textbf{CoreSet}                 & \multicolumn{1}{c|}{0.7461 $\pm$ 0.0035} & \multicolumn{1}{c|}{0.7950 $\pm$ 0.0208} & \multicolumn{1}{c|}{0.8065 $\pm$ 0.0070} & \multicolumn{1}{c|}{0.8114 $\pm$ 0.0072} & \multicolumn{1}{c|}{0.8295 $\pm$ 0.0022} & 0.8309 $\pm$ 0.0035 \\ 
\textbf{MCD-Entropy}             & \multicolumn{1}{c|}{0.7461 $\pm$ 0.0035} & \multicolumn{1}{c|}{0.7925 $\pm$ 0.0132} & \multicolumn{1}{c|}{0.8036 $\pm$ 0.0055} & \multicolumn{1}{c|}{0.8218 $\pm$ 0.0026} & \multicolumn{1}{c|}{0.8333 $\pm$ 0.0126} & 0.8424 $\pm$ 0.0127 \\ 
\textbf{MCD-BALD}                & \multicolumn{1}{c|}{0.7461 $\pm$ 0.0035} & \multicolumn{1}{c|}{0.7951 $\pm$ 0.0125} & \multicolumn{1}{c|}{0.8155 $\pm$ 0.0011} & \multicolumn{1}{c|}{0.8225 $\pm$ 0.0044} & \multicolumn{1}{c|}{0.8273 $\pm$ 0.0077} & 0.8344 $\pm$ 0.0130 \\ 
\textbf{AEIG}                    & \multicolumn{1}{c|}{0.7461 $\pm$ 0.0035} & \multicolumn{1}{c|}{0.7985 $\pm$ 0.0155} & \multicolumn{1}{c|}{0.8197 $\pm$ 0.0072} & \multicolumn{1}{c|}{0.8269 $\pm$ 0.0017} & \multicolumn{1}{c|}{0.8363 $\pm$ 0.0117} & 0.8468 $\pm$ 0.0024 \\ \hline \hline
\end{tabular}%
}
\caption{Comparison of the Random, Entropy, CoreSet, MCD-Entropy, MCD-BALD, and AEIG based active learning sampling methods for both the DR dataset We report the mean and std of evaluation metric across five different runs. Model performance with the entire training set is 0.8561.}
\label{tab:DR-Bench}
\end{table}

\begin{table}[]
\resizebox{\textwidth}{!}{%
\begin{tabular}{l|cccccc}
\hline \hline
\multirow{2}{*}{\textbf{Method}} &
  \multicolumn{6}{c|}{\textbf{Percentage of Labeled Sample}} \\ \cline{2-7} 
 &
  \multicolumn{1}{c|}{\textbf{10}} &
  \multicolumn{1}{c|}{\textbf{15.83}} &
  \multicolumn{1}{c|}{\textbf{21.67}} &
  \multicolumn{1}{c|}{\textbf{27.5}} &
  \multicolumn{1}{c|}{\textbf{33.33}} &
  \textbf{39.17} \\ \hline \hline
\textbf{Random} &
  \multicolumn{1}{c|}{0.9003 $\pm$ 0.0121} &
  \multicolumn{1}{c|}{0.9213 $\pm$ 0.0021} &
  \multicolumn{1}{c|}{0.9419 $\pm$ 0.0028} &
  \multicolumn{1}{c|}{0.9499 $\pm$ 0.0044} &
  \multicolumn{1}{c|}{0.9559 $\pm$ 0.0030} &
  0.9589 $\pm$ 0.0041 \\ 
\textbf{Entropy} &
  \multicolumn{1}{c|}{0.9003 $\pm$ 0.0121} &
  \multicolumn{1}{c|}{0.9385 $\pm$ 0.0026} &
  \multicolumn{1}{c|}{0.9567 $\pm$ 0.0072} &
  \multicolumn{1}{c|}{0.9649 $\pm$ 0.0039} &
  \multicolumn{1}{c|}{0.9714 $\pm$ 0.0023} &
  0.9755 $\pm$ 0.0034 \\ 
\textbf{CoreSet} &
  \multicolumn{1}{c|}{0.9003 $\pm$ 0.0121} &
  \multicolumn{1}{c|}{0.9426 $\pm$ 0.0028} &
  \multicolumn{1}{c|}{0.9561 $\pm$ 0.0030} &
  \multicolumn{1}{c|}{0.9642 $\pm$ 0.0011} &
  \multicolumn{1}{c|}{0.9703 $\pm$ 0.0015} &
  0.9745 $\pm$ 0.0014 \\ 
\textbf{MCD-Entropy} &
  \multicolumn{1}{c|}{0.9003 $\pm$ 0.0121} &
  \multicolumn{1}{c|}{0.9372 $\pm$ 0.0047} &
  \multicolumn{1}{c|}{0.9519 $\pm$ 0.0030} &
  \multicolumn{1}{c|}{0.9651 $\pm$ 0.0023} &
  \multicolumn{1}{c|}{0.9707 $\pm$ 0.0010} &
  0.9743 $\pm$ 0.0034 \\ 
\textbf{MCD-BALD} &
  \multicolumn{1}{c|}{0.9003 $\pm$ 0.0121} &
  \multicolumn{1}{c|}{0.9410 $\pm$ 0.0067} &
  \multicolumn{1}{c|}{0.9540 $\pm$ 0.0044} &
  \multicolumn{1}{c|}{0.9672 $\pm$ 0.0024} &
  \multicolumn{1}{c|}{0.9694 $\pm$ 0.0049} &
  0.9747 $\pm$ 0.0020 \\ 
\textbf{AEIG} &
  \multicolumn{1}{c|}{0.9003 $\pm$ 0.0121} &
  \multicolumn{1}{c|}{0.9439 $\pm$ 0.0040} &
  \multicolumn{1}{c|}{0.9577 $\pm$ 0.0028} &
  \multicolumn{1}{c|}{0.9681 $\pm$ 0.0046} &
  \multicolumn{1}{c|}{0.9735 $\pm$ 0.0022} &
  0.9753 $\pm$ 0.0018 \\ \hline \hline
\end{tabular}%
}
\caption{Comparison of the Random, Entropy, CoreSet, MCD-Entropy, MCD-BALD, and AEIG based active learning sampling methods for both the ISIC dataset We report the mean and std of evaluation metric across five different runs. Model performance with the entire training set is 0.9789.}
\label{tab:ISIC-Bench}
\end{table}

\subsection*{Implementation Details} 

\begin{figure*}[!t]
\centering
\includegraphics[width=1.00\textwidth]{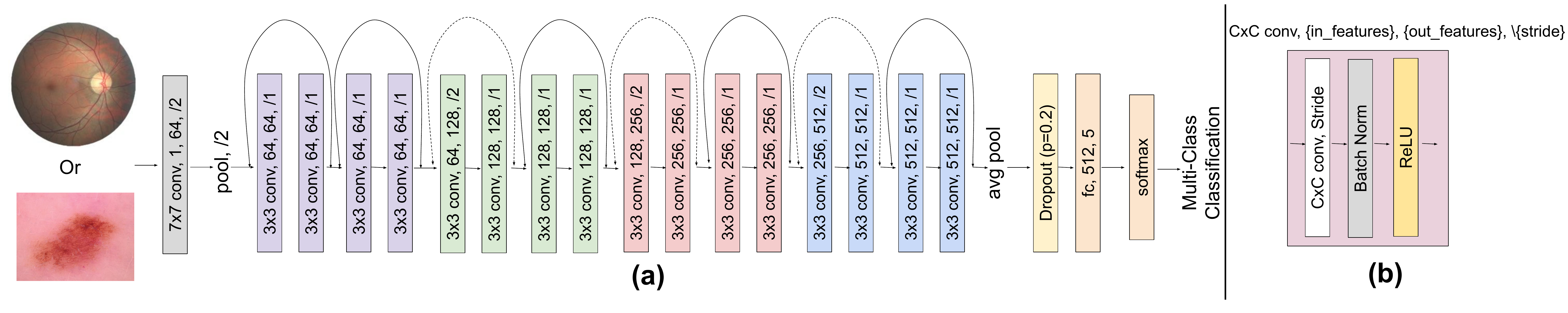}
\center \caption{(a) A 2D ResNet-18 architecture consists of a 7x7 convolutional unit, followed by 16 3x3 convolutional units, one dropout layer (p=0.2), and one fully connected layers. The dotted shortcuts increase dimensions. Colour fundus images (or dermoscopic images) were given as input to the network. (b) Each convolutional unit consists of one CxC convolutional layer with stride S, followed by Batch Normalization layer, and a ReLU layer. }
\label{fig:2DResNet18}
\end{figure*}

An ImageNet pre-trained 2D ResNet18 \cite{ResNet} architecture was used for the DR and the ISIC multi-class disease scale classification task. The network architecture is depicted in Fig.\ref{fig:2DResNet18}. A Dropout layer with p=0.2 is introduced before the fully connected (fc) layer. The network was trained to reduce the categorical cross entropy loss.  An Adam optimizer with a learning rate of $0.0005$ and a weight decay of $0.00001$ was used to train the network for a total of 100 epochs, and batch size of 64. The learning rate was decayed with a factor of $0.995$ after each epoch. All fundus images (DR) were resized to 512x512 size and normalized with mean subtraction and divide by std. All dermographic images (ISIC) were resized to 600x450 size and normalized with mean subtraction and divide by std. 
Random Horizontal Flip, Random Vertical Flip, and Random rotation in the range of 0-30, was applied as data augmentation on each image. The code was written in PyTorch and ran on Nvidia GeForce RTX 3090 GPU with 24GB memory. 

\bibliographystyle{splncs04}
\bibliography{biblio}

\end{document}